\documentclass[10pt,twocolumn]{ICCAS}
 
\usepackage{diagbox}
\usepackage{multirow}
\usepackage{url}  
\usepackage{booktabs}
\usepackage{hyperref}
\begin{document}

\title{Situation Aware Locomotion for Dual Mobile Cobots in Shared Environments}

\author{William Moraes${}^{1}$, Igor Nunes${}^{1}$, Ahilen Mazondo${}^{1}$, Sebastian Barcelona${}^{1}$, \\ Hiago Sodre${}^{1}$, Pablo Moraes${}^{1}$ and Ricardo B. Grando${}^{1*}$}

\affils{ ${}^{1}$Robotics and AI Lab, Technological University of Uruguay, \\
Rivera, Uruguay (ricardo.bedin@utec.edu.uy){\small${}^{*}$ Corresponding author}}


\abstract{This paper presents a situation aware locomotion framework for two mobile collaborative robots operating in shared industrial environments. The proposed method models situation awareness through perception, comprehension, and projection to support locomotion decisions. Robot pose, load state, manipulator state, shared zone occupancy, obstacle state, and predicted inter robot conflict were used to select safe locomotion actions. The framework was implemented in simulation and evaluated in simulated industrial scenarios designed to match a feasible 4 m by 4 m physical test area. The proposed method was compared with two other baselines over multiple trials and randomized seeds. The results show that the situation aware method achieved 100\% task success across all scenarios, while the independent and fixed priority baselines each achieved 33.3\% overall success. The proposed method eliminated shared zone conflicts and safety stops, maintained the largest average minimum inter robot distance, and completed the tasks with the lowest average completion time. These results indicate that situational awareness can improve the locomotion of dual robots by combining load state, manipulator state, reasoning about the shared zone, and prediction of short-horizon conflicts.
}


\maketitle


\section{Introduction}

Mobile collaborative robots are increasingly being considered for flexible industrial environments where tasks are performed in shared workspaces. In these environments, robot motion cannot be planned only as individual goal reaching. A robot may need to yield, wait, replan, or avoid entering a constrained area when another robot is already using the same space. These requirements become more relevant when mobile bases are combined with manipulators, because the manipulator posture or load condition can change the meaning of the robot state even when no active manipulation is being executed.

This work considers two LIMO Cobot robots, each composed of a LIMO mobile base and a myCobot 280 M5 manipulator. The LIMO Cobot platform integrates mobile robot navigation with a lightweight collaborative arm, while the myCobot 280 M5 provides a compact six degree of freedom manipulator for laboratory scale robotic experiments \cite{elephant_limo,elephant_mycobot}. Because the manipulator has a limited payload and reach, this study does not address full collaborative manipulation, grasping, or handover. Instead, the manipulator is used as part of the situation state. For example, a robot may be represented as loaded when the arm is in a loaded transport posture, while another robot may be represented as unloaded when the arm is in a safe locomotion posture.

The paper focuses on situation aware locomotion for dual mobile cobots. In this setting, the robots must reason about their own state, the state of the other robot, shared zone occupancy, obstacle conditions, and predicted motion conflicts. This formulation follows the situation awareness model proposed by Endsley, where decision making is organized around perception, comprehension, and projection \cite{endsley1995}. In the proposed method, perception is associated with robot pose, velocity, load state, manipulator state, shared zone geometry, and obstacle state. Comprehension is used to interpret which robot should proceed, yield, or recover. Projection is used to estimate whether the current motion command may lead to a shared zone conflict, unsafe proximity, or obstacle violation.

Two reference baselines are used to evaluate the proposed method. The first baseline is independent navigation, inspired by decoupled multi robot planning, where each robot follows its own goal directed path without shared situation interpretation \cite{lavalle1998}. The second baseline is fixed priority coordination, inspired by prioritized planning, where agents are assigned an order and lower priority agents yield to higher priority agents \cite{erdmann1987,bennewitz2001}. These baselines were selected because they represent two common simplifications in multi robot locomotion: navigation without coordination and coordination with a fixed rule.

\begin{figure*}[t]
    \centering
    \includegraphics[width=0.75\textwidth]{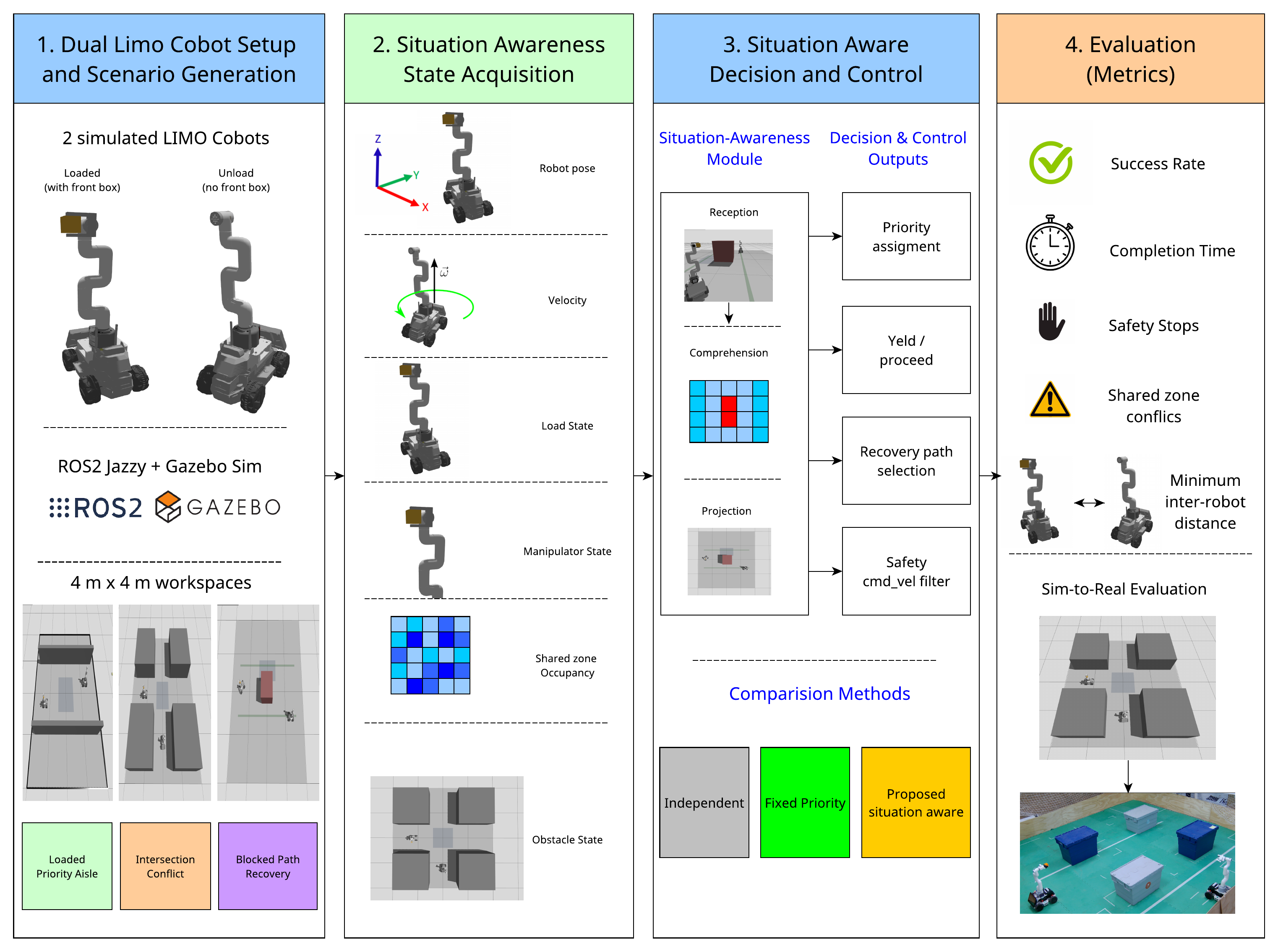}
    \caption{Overview of the proposed situation aware locomotion framework for dual mobile cobots. The framework integrates scenario generation, situation awareness state acquisition, situation aware decision and control, and evaluation metrics.}
    \label{fig:framework}
\end{figure*}

The methodology was implemented in ROS 2 Jazzy and Gazebo Sim. Three 4 m by 4 m scenarios were designed to preserve laboratory scale feasibility for future real robot deployment. The scenarios evaluate loaded robot priority, intersection conflict, and blocked path recovery. Each scenario was tested with the independent baseline, the fixed priority baseline, and the proposed situation aware method. A total of 900 trials were conducted using five randomized seeds.

Figure~\ref{fig:framework} summarizes the proposed framework. The system starts with two simulated LIMO Cobots and scenario generation. Robot pose, velocity, load state, manipulator state, shared zone occupancy, and obstacle state are then acquired as situation awareness inputs. These variables are processed through perception, comprehension, and projection layers. The resulting decision layer assigns priority, determines whether each robot should yield or proceed, selects a recovery path when needed, and applies a safety command filter before evaluating the trial. The final evaluation reports success rate, completion time, minimum inter robot distance, safety stops, and shared zone conflicts.

The contributions of this paper are summarized as follows:

\begin{itemize}
    \item A situation aware locomotion methodology is proposed for two mobile cobots operating in shared industrial environments.
    
    \item A manipulator state representation is introduced for locomotion level decision making, where the arm is used to indicate loaded, unloaded, waiting, or safe locomotion conditions.
    
    
    \item Three experimental scenarios are defined to evaluate loaded robot priority, shared intersection coordination, and blocked path recovery.
    
    \item The proposed method is compared with independent navigation and fixed priority baselines over 900 trials using success rate, completion time, minimum inter robot distance, safety stops, and shared zone conflicts.
\end{itemize}

The remainder of this paper is organized as follows. Section II reviews related work on situation awareness, independent and prioritized multi robot planning, and sim to real evaluation. Section III describes the proposed methodology, including the situation state representation, baselines, safety filter, and evaluation protocol. Section IV presents the experimental results and discussion. Section V concludes the paper and outlines future validation with physical LIMO Cobot robots. Open source repository: \url{https://github.com/ricardoGrando/limo_cobot_jazzy_sim}. A supplementary video is available at: \url{https://youtu.be/fMbDYO_ABzI}.
\vspace{-5mm}
\section{Related Work}

\subsection{Situation Awareness in Robotic Systems}

Situation awareness was formalized as the perception of elements in the environment, comprehension of their meaning, and projection of their status in the near future \cite{endsley1995}. Although the model was initially developed for human decision making in dynamic systems, it has also been applied to robotic systems because autonomous robots must interpret environmental conditions rather than only process raw sensor data. In mobile robotics, situation awareness can include robot pose, velocity, obstacle state, task state, and the state of other agents in the workspace.

For multi robot locomotion, situation awareness is useful because the meaning of a robot state depends on the shared context. A robot that is approaching an intersection may be safe when alone, but unsafe when another robot is approaching the same zone. Similarly, a mobile cobot in a loaded transport pose may require priority in a narrow aisle, even when the manipulator is not actively moving. In this paper, situation awareness is represented through robot pose, load state, manipulator state, shared zone occupancy, obstacle state, and predicted conflict.

\subsection{Independent and Decoupled Multi Robot Planning}

A common simplification in multi robot planning is to plan for each robot separately, then evaluate or resolve interactions afterward. LaValle and Hutchinson studied motion planning for multiple robots with independent goals and described planning formulations in which each robot follows its own objective while coordination is considered separately \cite{lavalle1998}. This family of approaches is related to decoupled multi robot planning, where the planning problem is reduced by avoiding full joint state planning.

In this work, the independent baseline was inspired by this decoupled planning idea. Each robot follows its own waypoint sequence toward its goal, without using shared situation awareness, load state, or manipulator state. This baseline was included to test whether goal directed locomotion alone is sufficient in shared industrial environments. It also provides a reference case for measuring shared zone conflicts and safety stops when coordination is absent.

\subsection{Prioritized Planning and Fixed Priority Coordination}

Prioritized planning is another common approach for coordinating multiple moving agents. Erdmann and Lozano Perez proposed an early formulation for planning with multiple moving objects by assigning a priority ordering and planning motions sequentially according to that order \cite{erdmann1987}. Later work by Bennewitz et al. applied prioritized planning to multi robot systems and showed how schedules can be optimized when robots are planned according to priority constraints \cite{bennewitz2001}.

The fixed priority baseline in this paper was inspired by prioritized planning. Robot 1 was always assigned priority whenever a potential shared zone conflict was detected, and Robot 2 was required to yield. This baseline represents a simple coordination method that can resolve some shared zone conflicts without modeling the broader situation. However, it does not adapt to load state, manipulator state, or blocked path conditions.

\subsection{Simulation and Sim to Real Evaluation}

Simulation is widely used in robotics because it allows robot algorithms to be evaluated before physical deployment. Gazebo was introduced as an open source multi robot simulator for three dimensional dynamic environments and sensor based robotic systems \cite{koenig2004}. In this work, Gazebo Sim was used with ROS 2 Jazzy. ROS 2 provides software libraries and tools for building robotic applications, while Gazebo Sim can be integrated with ROS 2 through the \texttt{ros\_gz} interface \cite{ros2docs,gazebo_docs}.

Sim to real transfer remains difficult because simulated and physical robots differ in sensing, actuation, dynamics, and environmental uncertainty. Prior work has described this difference as the reality gap and has reviewed approaches such as domain randomization, domain adaptation, imitation learning, and knowledge distillation \cite{zhao2020}. Tobin et al. showed that randomized simulation can support transfer to real world perception tasks by making the real world appear as one variation of the simulated distribution \cite{tobin2017}. In the present work, the method is not based on deep learning, but seed based perturbations were used to evaluate repeatability under small changes in initial robot pose.

\subsection{Research Gap and Contribution}

The reviewed literature provides established foundations for situation awareness, decoupled multi robot planning, prioritized planning, and simulation based evaluation. However, less attention has been given to lightweight mobile cobot teams where the manipulator is not used for manipulation, but still contributes to the interpretation of the robot state. In such systems, the manipulator posture can indicate whether the robot is loaded, waiting, or moving in a safe locomotion condition. This information can affect locomotion decisions even when no grasping or handover is performed.

This paper addresses that gap by proposing a situation aware locomotion methodology for two mobile cobots in shared industrial environments. The proposed method differs from independent planning because it uses shared zone state and predicted conflict. It differs from fixed priority planning because it uses load state, manipulator state, and obstacle state to adapt the locomotion behavior. The contribution is therefore not a new manipulator control method, but a sim to real evaluation framework in which manipulation related state is incorporated into locomotion level situation awareness.
\vspace{-5mm}
\section{Methodology}

\subsection{System Overview}

The proposed methodology was designed to evaluate situation aware locomotion with two mobile cobots in a shared industrial workspace. The system was implemented using two simulated LIMO Cobot robots. Each robot consisted of a LIMO mobile base and a myCobot 280 M5 manipulator. The experiments were implemented in ROS 2 Jazzy and Gazebo Sim. A kinematic base controller was used to execute planar motion commands, while wheel rotation was animated for visual consistency. A command safety filter was placed between the policy output and the simulator controller. Therefore, each method generated a requested velocity command, and the safety layer either accepted the command or replaced it with a zero velocity command when a predicted violation was detected.

Each scenario was limited to a 4 m by 4 m workspace. This workspace size was selected to match a feasible laboratory environment for future physical experiments with two LIMO Cobot platforms. Three scenarios were evaluated: loaded robot priority, intersection conflict, and blocked path recovery.

\subsection{Robot and Workspace State}

At each control cycle, the state of the dual robot system was represented as

\begin{equation}
S_t = \{R_1, R_2, V_1, V_2, L_1, L_2, A_1, A_2, Z, O, C\},
\end{equation}

where \(R_i = (x_i, y_i, \theta_i)\) is the planar pose of robot \(i\), \(V_i = (v_i, \omega_i)\) is its commanded velocity, \(L_i\) is its load state, \(A_i\) is its manipulator state, \(Z\) is the shared zone state, \(O\) is the obstacle state, and \(C\) is the predicted conflict state. The load state was defined as a binary variable, (1) if loaded and (0) not loaded.

Four manipulator states were used: safe locomotion pose, loaded transport pose, waiting pose, and sensing pose. These states were used to represent the task condition of the mobile cobot. The shared zone state was computed from the robot poses and predefined scenario geometry. For a rectangular shared zone with center \((x_z, y_z)\), length \(l_z\), and width \(w_z\), robot \(i\) was considered inside the zone when

\begin{equation}
|x_i - x_z| \leq \frac{l_z}{2}
\quad \text{and} \quad
|y_i - y_z| \leq \frac{w_z}{2}.
\end{equation}

The shared zone was considered occupied when at least one robot was inside the zone. A zone conflict was recorded when both robots occupied the shared zone at the same time or when the policy allowed both robots to enter the zone without a valid yielding decision.

\subsection{Situation Awareness Model}

The proposed method followed a three level situation awareness model composed of perception, comprehension, and projection.

\subsubsection{Perception Level}

At the perception level, the system collected robot pose, velocity command, manipulator state, load state, shared zone geometry, and obstacle geometry. In simulation, robot pose was obtained from Gazebo odometry. In a real robot implementation, the same state variables can be obtained from wheel odometry, inertial sensing, localization, range sensing, and robot to robot communication. The perception layer generated the following observation for each robot:

\begin{equation}
O_i(t) = \{R_i(t), V_i(t), L_i, A_i, G_i, P_i\},
\end{equation}

where \(G_i\) is the current goal and \(P_i\) is the active waypoint sequence.

\subsubsection{Comprehension Level}

At the comprehension level, the system interpreted the perceived state in relation to the scenario. Several conditions were evaluated: whether each robot was loaded or unloaded; whether each robot was inside or approaching the shared zone; whether the current goal required passage through a constrained area; whether a robot should proceed, yield, or move to a waiting point; whether the direct path was blocked and a recovery path was required.

The priority score for robot \(i\) was defined as

\begin{equation}
P_i = \alpha L_i + \beta Q_i - \gamma D_i - \delta C_i,
\end{equation}

where \(L_i\) is the load state, \(Q_i\) is the task urgency, \(D_i\) is the distance from the robot to the shared zone, and \(C_i\) is the predicted conflict cost. The weights \(\alpha\), \(\beta\), \(\gamma\), and \(\delta\) determine the relative contribution of each term. In the implemented policy, loaded state and conflict state were used as rule based terms, so the priority score was used as a compact representation of the decision logic rather than as a learned value function.

\subsubsection{Projection Level}

At the projection level, future poses were estimated using a short horizon unicycle model. For robot \(i\), the predicted pose after a time interval \(\Delta t\) was computed as

\begin{equation}
\hat{x}_i(t+\Delta t) =
x_i(t) + v_i \cos(\theta_i) \Delta t,
\end{equation}

\begin{equation}
\hat{y}_i(t+\Delta t) =
y_i(t) + v_i \sin(\theta_i) \Delta t,
\end{equation}

\begin{equation}
\hat{\theta}_i(t+\Delta t) =
\theta_i(t) + \omega_i \Delta t.
\end{equation}

A predicted inter robot conflict was detected when

\begin{equation}
d(\hat{R}_1, \hat{R}_2) < d_{\min},
\end{equation}

where \(d(\hat{R}_1, \hat{R}_2)\) is the predicted Euclidean distance between both robots and \(d_{\min}\) is the minimum allowed separation. The same prediction horizon was also used to detect future intersection with obstacle boxes and restricted shared zone conditions.

\subsection{Low Level Motion Command}

Each robot was commanded using a simple waypoint tracking controller. Given the current robot pose \((x_i, y_i, \theta_i)\) and the active waypoint \((x_g, y_g)\), the distance and heading error were computed as

\begin{equation}
e_d = \sqrt{(x_g - x_i)^2 + (y_g - y_i)^2},
\end{equation}

\begin{equation}
e_{\theta} =
\mathrm{atan2}(y_g-y_i, x_g-x_i) - \theta_i.
\end{equation}

The requested linear and angular velocities were computed as

\begin{equation}
v_i = \mathrm{clip}(k_v e_d, 0, v_{\max}),
\end{equation}

\begin{equation}
\omega_i = \mathrm{clip}(k_{\omega} e_{\theta}, -\omega_{\max}, \omega_{\max}).
\end{equation}

When the heading error was above a predefined threshold, the linear velocity was reduced so that the robot first aligned toward the waypoint. A waypoint was considered reached when the robot was within a tolerance radius of the waypoint. After that, the next waypoint in the sequence was activated.

\begin{table*}[t]
\centering
\caption{Scenario level results. Continuous metrics are reported as mean $\pm$ standard deviation across 100 trials. Best values in each scenario and metric are shown in bold.}
\label{tab:scenario_results}
\renewcommand{\arraystretch}{1.08}
\setlength{\arrayrulewidth}{0.6pt}
\resizebox{\textwidth}{!}{
\begin{tabular}{llcccc}
\hline
\textbf{Scenario} & \textbf{Method} & \textbf{Success} & \textbf{Time (s)} & \textbf{Stops} & \textbf{Conflicts} \\
\hline
S1 & Independent & \textbf{100.0\%} & \textbf{14.36 $\pm$ 0.05} & \textbf{0.00} & 1.00 $\pm$ 0.00 \\
S1 & Fixed priority & 0.0\% & 55.01 $\pm$ 0.03 & \textbf{0.00} & \textbf{0.00} \\
S1 & Situation aware & \textbf{100.0\%} & 30.58 $\pm$ 0.14 & \textbf{0.00} & \textbf{0.00} \\
\hline
S2 & Independent & 0.0\% & 50.02 $\pm$ 0.04 & 159.29 $\pm$ 1.44 & 1.00 $\pm$ 0.00 \\
S2 & Fixed priority & \textbf{100.0\%} & \textbf{21.36 $\pm$ 0.09} & \textbf{0.00} & \textbf{0.00} \\
S2 & Situation aware & \textbf{100.0\%} & \textbf{21.36 $\pm$ 0.09} & \textbf{0.00} & \textbf{0.00} \\
\hline
S3 & Independent & 0.0\% & 70.02 $\pm$ 0.04 & 117.39 $\pm$ 0.84 & \textbf{0.00} \\
S3 & Fixed priority & 0.0\% & 70.01 $\pm$ 0.03 & 112.48 $\pm$ 1.81 & \textbf{0.00} \\
S3 & Situation aware & \textbf{100.0\%} & \textbf{19.79 $\pm$ 0.12} & \textbf{0.00} & \textbf{0.00} \\
\hline
\end{tabular}
}
\end{table*}

\subsection{Baseline Methods}


\subsubsection{Independent Navigation Baseline}

The independent baseline was inspired by decoupled multi robot planning, where each robot follows its own path or goal and coordination is not included in the main decision layer \cite{lavalle1998}. In the present implementation, each robot followed its own waypoint sequence toward the goal. The robot did not use the load state, manipulator state, shared zone state, or the task state of the other robot.

For robot \(i\), the path used by the independent baseline was defined as

\begin{equation}
P_i = P_i^{default}.
\end{equation}

No priority was assigned, and no waiting point was selected. The safety filter was still active for this baseline, so unsafe commands could be rejected. However, the independent baseline did not include any coordinated recovery behavior after a command was stopped. This baseline was included to evaluate whether individual goal directed locomotion was sufficient in shared workspaces.

\subsubsection{Fixed Priority Baseline}

The fixed priority baseline was inspired by prioritized planning, where agents are assigned an ordering and lower priority agents must avoid or yield to higher priority agents \cite{erdmann1987,bennewitz2001}. In this implementation, Robot 1 was always assigned priority when a predicted conflict or shared zone conflict was detected. Robot 2 yielded until the conflict condition was cleared.

The fixed priority rule was defined as

\begin{equation}
\pi =
\begin{cases}
R_1, & \text{if a conflict is predicted},\\
\emptyset, & \text{otherwise}.
\end{cases}
\end{equation}

When Robot 1 had priority, Robot 2 was commanded to stop or move to a predefined waiting point outside the shared zone. Robot 1 continued toward its active waypoint. When no conflict was predicted, both robots followed their default waypoint sequences. This baseline was included because it represents a simple and common coordination rule. It can solve some intersection conflicts, but it does not use load state, manipulator state, or blocked path recovery.

\subsection{Experimental Scenarios}

Three scenarios were used to evaluate different aspects of the methodology. Each scenario was defined in a 4 m by 4 m workspace. S1 Loaded priority: tests whether a loaded robot receives priority in a narrow shared aisle while avoiding zone conflicts. S2 Intersection: tests whether two robots approaching a shared intersection can coordinate yielding. S3 Blocked path: tests whether the method can detect a blocked route and activate a recovery path. The scenarios were replicated in the real world ensuring the same dimensions and features.




\vspace{-5mm}
\section{Results and Discussion}

\begin{figure*}[t]
    \centering
    \includegraphics[width=0.86\textwidth]{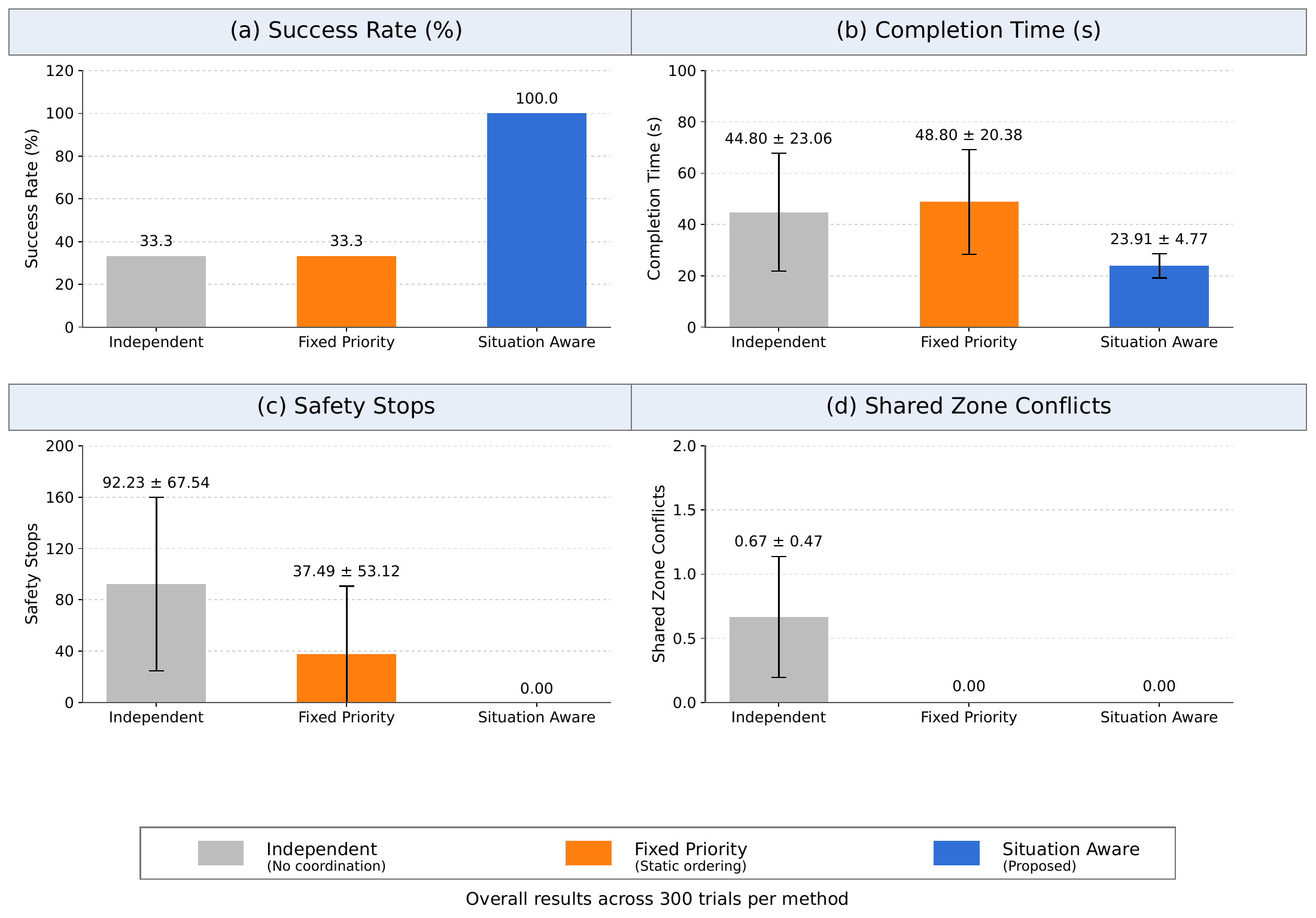}
    \caption{Overall performance comparison across all scenarios and methods. The results summarize success rate, completion time, safety stops, and shared zone conflicts over 300 trials per method. The situation aware method achieved full task success while eliminating safety stops and shared zone conflicts.}
    \label{fig:results_summary}
\end{figure*}

Figure~\ref{fig:results_summary} summarizes the overall results. The situation aware method achieved 100.0\% success across 300 trials, while the independent and fixed priority baselines each achieved 33.3\%. The proposed method also obtained the lowest average completion time and eliminated safety stops and shared zone conflicts.

Table~\ref{tab:scenario_results} reports the scenario level results. In S1, the independent baseline reached the goals quickly, but it produced one shared zone conflict in every trial. The fixed priority baseline avoided conflicts, but failed by timeout. The situation aware method completed all trials without shared zone conflicts, showing that load and manipulator state can support locomotion level priority assignment.

In S2, the independent baseline failed because both robots attempted to enter the shared intersection without yielding. Fixed priority and situation awareness achieved the same success rate because this scenario required only a simple priority decision. In S3, both baselines failed because they followed the default path and did not activate recovery. The situation aware method succeeded in all trials by interpreting the obstacle state and selecting the recovery path. This result shows the added value of situation awareness beyond fixed priority coordination. Overall, the results show that independent navigation is unsafe in shared zones, fixed priority is effective only for simple conflicts, and situation awareness provides reliable behavior across all scenarios.

\vspace{-5mm}
\section{Conclusion}

This paper presented a situation aware locomotion framework for two mobile cobots in shared industrial environments, both simulation and real world simulated scenarios. The manipulator was used as a situation state indicator rather than for active manipulation. Across 900 simulated trials, the proposed method achieved 100.0\% success, while independent navigation and fixed priority each achieved 33.3\%. The proposed method also eliminated safety stops and shared zone conflicts. Future work will compare simulation and real robot performance under controlled perturbations.

\vspace{-5mm}
\section{Acknowledgements}

The authors of this work would like to thank the Technological University of Uruguay and the Laboratory of Robotics and AI for the support in this work.

\vspace{-5mm}

\bibliographystyle{./bibliography/IEEEtran}
\bibliography{./bibliography/IEEEabrv,./bibliography/main}

\end{document}